\documentclass[11pt,a4paper]{article}
\pdfoutput=1
\usepackage[hyperref]{emnlp2020}
\usepackage{times}
\usepackage{latexsym}

\usepackage{microtype}

\aclfinalcopy % Uncomment this line for the final submission
\def\aclpaperid{2} %  Enter the acl Paper ID here

\usepackage{amsmath}
\usepackage{amssymb}

\newcommand{\bb}{\fontseries{b}\selectfont}
\newcommand{\comment}[1]{}
\DeclareMathOperator{\BigO}{O}
\newcommand{\R}{\mathbb{R}}
\usepackage{hhline}
\usepackage{ctable}

\title{Leader: Prefixing a Length for Faster Word Vector Serialization}

\author{Brian Lester \\
  Independent \\
  \texttt{blester125@gmail.com} \\
}

\date{}

\begin{document}

\maketitle
\begin{abstract}
Two competing file formats have become the de facto standards for distributing pre-trained word embeddings. Both are named after the most popular pre-trained embeddings that are distributed in that format. The GloVe format is an entirely text based format that suffers from huge file sizes and slow reads, and the word2vec format is a smaller binary format that mixes a textual representation of words with a binary representation of the vectors themselves. Both formats have problems that we solve with a new format we call the Leader format. We include a word length prefix for faster reads while maintaining the smaller file size a binary format offers. We also created a minimalist library to facilitate the reading and writing of various word vector formats, as well as tools for converting pre-trained embeddings to our new Leader format.

\end{abstract}

\section{Introduction}

Word vectors, often called word embeddings or word representations, are a vector space model where the semantic meaning of words are encoded as a low dimensional vector of real numbers. Effective word vectors are ones where vector similarity, computed with an inner product, is correlated with the similarity between the words they represent. There are many different algorithms for the creation of word vectors, but we will instead focus on storage and serialization, which is agnostic to the method used to create them.

Conceptually, these word vectors are a map from a word to a vector, but are often stored in two associated arrays. One is a list of words $W = [w_1, w_2, \dots, w_n]$ and the other is a matrix $V \in \R^{n \text{ x } v}$ formed by stacking the array of word vectors. These are aligned so that the word at index $i$, $W_i$, is represented by the vector $V_i \in \R^v$. Often a hash-table is used to map the words in $W$ to integer indices while providing $\BigO(1)$ lookup.

There are several options for serializing these data structures. They all make slightly different choices with different trade-offs in terms of file size, read time, and ease of implementation. We summarize common formats here.

GloVe \cite{penningtonGloVeGlobalVectors2014} is a pure text format where each line represents a (word, vector) tuple. The line itself is made of the word followed by all the elements of the vector, represented as text, with spaces as delimiters. The well-known GloVe embeddings are distributed in this format.

The word2vec \cite{mikolovEfficientEstimationWord2013, mikolovDistributedRepresentationsWords2013} software package introduced two formats, one text-based and the other binary. The text version was the basis for the GloVe format, with the only difference being that the word2vec text format has a header line with two integers (represented as text). These two integers are the size of the vocabulary and the size of the vectors respectively. Both FastText \cite{bojanowskiEnrichingWordVectors2017} and numberbatch \cite{speer2017conceptnet} use this format as a distribution medium.

The word2vec binary format is a mix of text and binary representations. The file starts with a text based header, again, two integers, representing the vocabulary and vector size. Following the header are (word, vector) tuples. These tuples are represented by: the word as text, a space, and then the elements of the vector in binary as float32s. The Google News pre-trained vectors are the most popular vectors that use this format.

A third less common format was used for the Senna embeddings \cite{JMLR:v12:collobert11a}. The Senna embeddings use a two-file format which is a more literal representation of word vectors as two associated arrays, one containing words and the other vectors. One file contains the words, one word per line, and the other has the vector list, one  vector per line, represented via text like GloVe. These files are aligned so that the word on line $i$ in the word file was represented by the vector on line $i$ in the vector file.

Problems with each of these formats has led us to introduce a new vector serialization format we call Leader, as well as a small, efficient library for working with these word vectors.

\section{Problems with Word Vector Serialization}

Here we outline specific problems with each of these formats, as well as general problems in the word vector ecosystem that have led to the creation of both our format and our library.

The formats with the most glaring problems are the GloVe and word2vec text formats. Their biggest advantage is that writing custom code for reading is very straight forward, but this is a small consolation when faced with the massive file size and incredibly slow load times caused by the text representation. The formats share the same problems because they are same format---save the header in the word2vec text format.

The word2vec binary format has issues of its own. The first is that binary formats are intimidating and assumed to be complex. This causes most researchers to either shy away from the format altogether, or to turn to a large external library to read it for them. The second problem is that the word and vector are delimited by a space, and we do not know the length of the word a priori. This means we need to iterate over the word, character by character, to find where the vector begins. When the vocabulary is massive and the words start to get longer, this can cause computational slowdown in a language like Python. Another problem with this format is that there is no formal definition---the only reference for the format is the files produced by the original implementation. For example, the lack of an official decision on the endianness used in the vector representation can cause problems when vectors trained on a machine using one endian format are used on a machine that uses the other endian format.

The Senna format is plagued by the same large file and slow read problems that GloVe has because it is also a text based format. Plus, it adds the extra bookkeeping of having to track two files.

Beyond problems with specific formats, there are also general problems in the tooling built around these word vectors. The first is that there is no small dedicated library for working with these files. Instead, researchers bring in large libraries like Gensim \cite{rehurek_lrec} or Mead-Baseline \cite{presselBaselineLibraryRapid2018} when reading a word vector file is all that is needed. In addition to the problem of pulling in a lot of extra code that will never be used, the majority of these libraries return specialized objects designed for their own downstream use of the word vectors. These objects are often superfluous additions and frustrating to deal with when the word vectors themselves are the goal.

The second problem is that perceived simplicity of the text formats, coupled with the heavy price of using a large library, has lead many to roll their own implementation of vector reading. This introduces subtle bugs and inconsistencies between projects. An example of an easy-to-miss bug is that in several of the pre-trained GloVe embeddings there are characters that are part of a word, but will cause the word to be broken in two when using Python's \texttt{.split()} method. Multiple re-implementations of the reader also leads to inconsistencies in the handling of edge cases. In the GloVe-840B pre-trained embedding, there is a duplicated word and different readers will make different policy decisions about if they should use the vector from the first or second occurrence of the word.

A third problem is one of community and communication surrounding how formats are named. The current nomenclature uses the name of the training algorithm that was introduced at the same time as the serialization format. Not only does this entangle the idea of vector training with serialization, it also leads to confusing questions, like ``Why do FastText vectors use the word2vec text format?'' The name of the format should tell us something about how the underlying data is structured, not just tell us who first made the format.

To combat the problems inherent to different formats we introduce a new format, Leader. We also release a small dedicated library for working with these files to help stem the community problems.

\section{The Leader Format}

To remedy the shortcomings of the various vector formats above, we introduce a new vector serialization scheme we call the Leader format. Leader is a fully binary format. It begins with a header made up of a 3-tuple of little-endian, unsigned, long longs. The members of this tuple are 1) a magic number, used to make sure we are operating on a valid Leader format file, 2) the number of types in the vocabulary and 3) the size of of word vectors. Following the header comes (size, word, vector) tuples for each word in the vocabulary. The size is the length of this particular word, encoded as utf-8 bytes rather than Unicode codepoints. The word is encoded as utf-8 bytes. The vector component of the tuple is the elements of the vector encoded as little-endian float32s (4 bytes). By including the length of the word in each tuple, we avoid the manual iteration over characters resulting in faster read times at the cost of just 3 bytes per word.

The Leader format is not a panacea; however, there are some difficulties that come with it. One difficulty is that, as a binary format, it is more difficult to get information from the file by hand. For example, to get the vocabulary size from a GloVe file, one just needs to count the lines in the file with \texttt{wc -l file}. In the word2vec formats, one can examine header with \texttt{head -n 1 file}. This information is available in the Leader header, too, it is just more complicated to extract it with \texttt{od -l -N 24 --endian=little file}.

\section{Benchmarks}
\label{sec:benchmarks}

\begin{table}[]
    \centering
    \begin{tabular}{l | r r} 
        Pre-trained & Vocabulary Size & Vector Size\\
        \hhline{=|==}
        GloVe 6B    &  400,000 & 100 \\
        GloVe 27B   & 1,193,514 & 200 \\
        GloVe 42B   & 1,917,494 & 300 \\
        GloVe $\text{840B}^*$ & 2,196,017 & 300 \\
        FastText Wiki & 999,994 & 300 \\
        FastText Crawl & 1,999,995 & 300 \\
        Google News & 3,000,000 & 300 \\
    \end{tabular}
    \caption{
        Statistics about the various pre-trained word embeddings used in our benchmarks. The GloVe 840B embedding actually has a duplicated word in it. For fairness in comparisons, we removed this duplicate so the real vocabulary size was $2,196,016$.
    }
    \label{tab:pre-trained-stats}
\end{table}

\begin{table}[t]
    \centering
    \begin{tabular}{l l | r}
        Pre-trained & Format  & File Size \\
        \hhline{==|=}
        GloVe 6b    & GloVe     & 0.3233 GiB \\
        %            & w2v Text  & 0.3233 GiB \\
                    & w2v       & 0.1521 GiB \\
                    & Leader & 0.1533 GiB \\
        \hline
        GloVe 27B   & GloVe     & 1.9163 GiB \\
        %            & w2v Text  & 1.9163 GiB \\
                    & w2v       & 0.9019 GiB \\
                    & Leader & 0.9053 GiB \\
        \hline
        GloVe 42B   & GloVe     & 4.6799 GiB \\
        %            & w2v Text  & 4.6799 GiB \\
                    & w2v       & 2.1594 GiB \\
                    & Leader & 2.1647 GiB \\
        \hline
        GloVe 840B  & GloVe     & 5.2585 GiB \\
        %            & w2v Text  & 5.2585 GiB \\
                    & w2v       & 2.4726 GiB \\
                    & Leader & 2.4788 GiB \\
        \hline
        FastText Wiki & GloVe     & 2.1039 GiB\\
        %              & w2v Text  & 2.1039 GiB \\
                      & w2v       & 1.1258 GiB \\
                      & Leader & 1.1286 GiB \\
        \hline
        FastText Crawl & GloVe     & 4.2046 GiB \\
        %               & w2v Text  & 4.2046 GiB \\
                       & w2v       & 2.2518 GiB \\
                       & Leader & 2.2574 GiB \\
        \hline
        Google News & GloVe     & 10.0271 GiB \\
        %            & w2v Text  & 10.0271 GiB \\
                    & w2v       &  3.3940 GiB \\
                    & Leader &  3.4025 GiB \\
    \end{tabular}
    \caption{
        The size on disk of pre-trained embeddings in different vector formats. The text based formats result in a much bigger file. The word2vec text format has been omitted because it only differs from the GloVe format by the size of the header---which is only 15 bytes at the most for the GloVe-840B embedding. Our Leader format is only slightly bigger than the word2vec format due to the 3 extra bytes used to store the length of the word rather than a space to delimit word and vector.
    }
    \label{tab:pre-trained-file-sizes}
\end{table}

\begin{table}[!t]
    \centering
    \begin{tabular}{l l | r r} 
        %& & \multicolumn{2}{c}{\enspace Read Time} \\
        %\cline{3-4}
        Pre-trained & Format & mean & std \\
        \hhline{==|==}
        GloVe 6B    & GloVe     & 11.26  & 0.29 \\
                    & w2v Text  & 11.15  & 0.10 \\
                    & w2v       &  1.97  & 0.12 \\
                    & Leader &  \bb{1.49} & 0.05 \\
        \hline
        GloVe 27B   & GloVe     & 61.20  & 0.32 \\
                    & w2v Text  & 62.06  & 1.03 \\
                    & w2v       &  6.97  & 0.09 \\
                    & Leader &  \bb{5.52} & 0.04 \\
        \hline
        GloVe 42B   & GloVe     & 136.27 & 0.16 \\
                    & w2v Text  & 138.53 & 2.30 \\
                    & w2v       &  11.22 & 0.28 \\
                    & Leader &  \bb{9.21} & 0.49 \\
        \hline
        GloVe 840B  & GloVe     & 154.08 & 0.23 \\
                    & w2v Text  & 154.12 & 0.33 \\
                    & w2v       &  12.92 & 0.35 \\
                    & Leader &  \bb{10.52} & 0.48 \\
        \hline
        FastText Wiki & GloVe & 67.10 & 0.18 \\
                      & w2v Text & 66.97 & 0.08 \\
                      & w2v & 5.82 & 0.19 \\
                      & Leader & \bb{4.76} & 0.25 \\
        \hline
        FastText Crawl & GloVe & 132.66 & 0.43 \\
                       & w2v Text & 136.86 & 2.14 \\
                       & w2v & 11.53 & 0.15 \\
                       & Leader & \bb{9.87} & 0.25 \\
        \hline
        Google News & GloVe     & 372.18 & 1.86 \\
                    & w2v Text  & 357.26 & 7.16 \\
                    & w2v       &  19.41 & 0.15 \\
                    & Leader &  \bb{14.82} & 0.06 \\
    \end{tabular}
    \caption{
        Time taken to read different pre-trained embeddings in different formats reported as the mean and standard deviation over 5 runs. The bold entries represent a statistically significant difference in speed. Benchmarks were preformed on a Intel(R) Core(TM) i7-7700HQ CPU @ 2.80GHz with 8 cores, 6144 KB cache, and 32819844 KB RAM using Python 3.6.10 \cite{10.5555/1593511} and Numpy \cite{oliphant2006guide,van2011numpy} 1.18.5 installed via Anaconda \cite{anaconda2016} 4.8.3 and word-vectors 4.0.0 from PyPI.
    }
    \label{tab:pre-trained-read-times}
\end{table}

To test the effectiveness of our file format, we took several pre-trained word embeddings and convert them into each format. These pre-trained vectors include GloVe vectors trained on various data sources, FastText pre-trained vectors \cite{mikolovAdvancesPreTrainingDistributed2018}, and word2vec vectors trained on Google News data.

We then compared each format in terms of both file size and read times when using our library that we discuss in Section \ref{sec:library}. A summary of the pre-trained embeddings used is found in Table \ref{tab:pre-trained-stats}. The file size of the different formats is shown in Table \ref{tab:pre-trained-file-sizes}. The binary formats give massive reductions in file sizes. The word2vec binary is the smallest, with our Leader format being only slightly bigger. The time taken to ingest the entire file can be found in Table \ref{tab:pre-trained-read-times}. Our Leader format is the fastest across the board. All benchmarks were done on the same machine and the page cache, dentrices, and inodes were reset between each read.

\section{Our Library}
\label{sec:library}

\begin{table}[]
    \centering
    \begin{tabular}{ll|rrr}
        %& & \multicolumn{3}{c}{Read Time} \\
        %\cline{3-5}
        Pre-trained & Format & mean         & std         & $\Delta$        \\
        \hhline{==|===}
        GV-6B       & Text   & 37.58        & 1.03        & \bb{70.33\%}  \\
                    & w2v    & 2.50         & 0.08        & \bb{21.20\%}  \\
        \hline
        GV-27B      & Text   & 210.47       & 6.70        & \bb{70.51\%}  \\
                    & w2v    & 7.52         & 0.31        &  \bb{7.31\%}  \\
        \hline
        GV-42B      & Text   & 484.95       & 1.77        & \bb{71.43\%}  \\
                    & w2v    & 12.93        & 0.17        & \bb{13.23\%}  \\
        \hline
        GV-840B     & Text   & 554.43       & 5.61        & \bb{72.20\%}  \\
                    & w2v    & 14.90        & 0.12        & \bb{13.29\%}  \\
        \hline
        FT Wiki     & Text   & 243.63       & 4.11        & \bb{71.76\%}  \\
                    & w2v    & 6.85         & 0.22        & \bb{15.04\%}  \\
        \hline
        FT Crawl    & Text   & 484.71       & 5.40        & \bb{71.76\%}  \\
                    & w2v    & 13.57        & 0.35        & \bb{15.03\%}  \\
        \hline
        GN          & Text   & 782.46       & 61.85       & \bb{54.34\%}  \\           
                    & w2v    & 21.54        & 0.81        &  \bb{9.89\%}  \\
    \end{tabular}
    \caption{
        Time taken to read files with Gensim and the percentage improvement using library. Our library improves read time of the text-based word2vec format by $68.90\%$ on average and improved the binary format by $12.79\%$. Times are reported over 5 runs, and bold entries are a statistically significant improvement. Benchmarks were performed on the same machine outlined in Table \ref{tab:pre-trained-read-times}. We used Gensim 3.8.3 from Anaconda.
    }
    \label{tab:gensim-bakeoff}
\end{table}

To help drive adoption of the better, binary formats available, as well as to provide a useful and consistent tool to the community, we also introduce a Python library, \textit{word-vectors}, for reading, writing, and converting between these formats. Our library is able to read and write all formats discussed above, except for Senna. The two-file approach would have caused a divergence in the API. Plus, the Senna format can be converted to GloVe with \texttt{paste -d" " word-file vector-file}.

We include a robust sniffing function that can automatically deduce the format of saved vectors. Our library is small, containing only core word vector I/O functionality, to ensure that bringing it in as a dependency is as painless as possible.

Our library is also performant. We compared our library to the popular Gensim library. Using the same pre-trained embeddings as before, in the two formats that Gensim supports (word2vec text and binary), we time how long it takes to read the whole file. Our benchmarks in Table \ref{tab:gensim-bakeoff} show that our library is much faster than Gensim across all pre-trained embeddings in all formats.

Our library also includes functionality to filter by a pre-defined vocabulary while reading so one does not have to materialize the entire pre-trained embedding matrix in memory. It is also well tested, using property based testing and automated CI/CD to ensure a consistent experience across supported operating systems and Python versions. 

\section{Conclusion}

The available word vector serialization formats have been plagued by efficiency and speed issues that have not been addressed due to the laissez-faire approach the community has taken on the standardization of word vector formats and tooling. We have outlined specific shortcomings in common formats, as well as problems in ecosystem surrounding word vectors. We have introduced the Leader format to solve the first set of problems and an open-source library for the second. Regardless of the traction our Leader format gets, our library will help the community migrate from the woefully inadequate text formats to superior binary ones.

\bibliography{emnlp2020}

\begin{thebibliography}{14}
\expandafter\ifx\csname natexlab\endcsname\relax\def\natexlab#1{#1}\fi

\bibitem[{Anaconda(2016)}]{anaconda2016}
Anaconda. 2016.
\newblock \href {https://anaconda.com/} {{A}naconda {S}oftware {D}istribution}.

\bibitem[{Bojanowski et~al.(2017)Bojanowski, Grave, Joulin, and
  Mikolov}]{bojanowskiEnrichingWordVectors2017}
Piotr Bojanowski, Edouard Grave, Armand Joulin, and Tomas Mikolov. 2017.
\newblock \href {https://doi.org/10.1162/tacl_a_00051} {{E}nriching {{Word
  Vectors}} with {{Subword Information}}}.
\newblock \emph{Transactions of the Association for Computational Linguistics},
  5:135--146.

\bibitem[{Collobert et~al.(2011)Collobert, Weston, Bottou, Karlen, Kavukcuoglu,
  and Kuksa}]{JMLR:v12:collobert11a}
Ronan Collobert, Jason Weston, L{\'e}on Bottou, Michael Karlen, Koray
  Kavukcuoglu, and Pavel Kuksa. 2011.
\newblock {N}atural {L}anguage {P}rocessing ({A}lmost) from {S}cratch.
\newblock \emph{Journal of Machine Learning Research}, 12(76):2493--2537.

\bibitem[{Mikolov et~al.(2013{\natexlab{a}})Mikolov, Chen, Corrado, and
  Dean}]{mikolovEfficientEstimationWord2013}
Tomas Mikolov, Kai Chen, Greg~S. Corrado, and Jeffrey Dean. 2013{\natexlab{a}}.
\newblock {E}fficient {{Estimation}} of {{Word Representations}} in {{Vector
  Space}}.
\newblock \emph{In Proceedings of Workshop at ICLR}.

\bibitem[{Mikolov et~al.(2018)Mikolov, Grave, Bojanowski, Puhrsch, and
  Joulin}]{mikolovAdvancesPreTrainingDistributed2018}
Tomas Mikolov, Edouard Grave, Piotr Bojanowski, Christian Puhrsch, and Armand
  Joulin. 2018.
\newblock {A}dvances in {{Pre}}-{{Training Distributed Word Representations}}.
\newblock In \emph{Proceedings of the {{Eleventh International Conference}} on
  {{Language Resources}} and {{Evaluation}} ({{LREC}} 2018)}, {Miyazaki,
  Japan}. {European Language Resources Association (ELRA)}.

\bibitem[{Mikolov et~al.(2013{\natexlab{b}})Mikolov, Sutskever, Chen, Corrado,
  and Dean}]{mikolovDistributedRepresentationsWords2013}
Tomas Mikolov, Ilya Sutskever, Kai Chen, Greg~S Corrado, and Jeff Dean.
  2013{\natexlab{b}}.
\newblock {D}istributed {{Representations}} of {{Words}} and {{Phrases}} and
  their {{Compositionality}}.
\newblock In C.~J.~C. Burges, L.~Bottou, M.~Welling, Z.~Ghahramani, and K.~Q.
  Weinberger, editors, \emph{Advances in {{Neural Information Processing
  Systems}} 26}, pages 3111--3119. {Curran Associates, Inc.}

\bibitem[{Oliphant(2006)}]{oliphant2006guide}
Travis~E Oliphant. 2006.
\newblock \emph{{A} {G}uide to {N}um{P}y}, volume~1.
\newblock Trelgol Publishing USA.

\bibitem[{Pennington et~al.(2014)Pennington, Socher, and
  Manning}]{penningtonGloVeGlobalVectors2014}
Jeffrey Pennington, Richard Socher, and Christopher Manning. 2014.
\newblock \href {https://doi.org/10.3115/v1/D14-1162} {{{GloVe}}: {{Global
  Vectors}} for {{Word Representation}}}.
\newblock In \emph{Proceedings of the 2014 {{Conference}} on {{Empirical
  Methods}} in {{Natural Language Processing}} ({{EMNLP}})}, pages 1532--1543,
  {Doha, Qatar}. {Association for Computational Linguistics}.

\bibitem[{Pressel et~al.(2018)Pressel, Ray~Choudhury, Lester, Zhao, and
  Barta}]{presselBaselineLibraryRapid2018}
Daniel Pressel, Sagnik Ray~Choudhury, Brian Lester, Yanjie Zhao, and Matt
  Barta. 2018.
\newblock \href {https://doi.org/10.18653/v1/W18-2506} {Baseline: {{A Library}}
  for {{Rapid Modeling}}, {{Experimentation}} and {{Development}} of {{Deep
  Learning Algorithms}} {T}argeting {{NLP}}}.
\newblock In \emph{Proceedings of {{Workshop}} for {{NLP Open Source Software}}
  ({{NLP}}-{{OSS}})}, pages 34--40, {Melbourne, Australia}. {Association for
  Computational Linguistics}.

\bibitem[{{\v R}eh{\r{u}}{\v r}ek and Sojka(2010)}]{rehurek_lrec}
Radim {\v R}eh{\r{u}}{\v r}ek and Petr Sojka. 2010.
\newblock {S}oftware {F}ramework for {T}opic {M}odelling with {L}arge
  {C}orpora.
\newblock In \emph{Proceedings of the {{LREC}} 2010 Workshop on New Challenges
  for {{NLP}} Frameworks}, pages 45--50, {Valletta, Malta}. {ELRA}.

\bibitem[{Speer et~al.(2017)Speer, Chin, and Havasi}]{speer2017conceptnet}
Robyn Speer, Joshua Chin, and Catherine Havasi. 2017.
\newblock \href {http://aaai.org/ocs/index.php/AAAI/AAAI17/paper/view/14972}
  {{ConceptNet} 5.5: {A}n {O}pen {M}ultilingual {G}raph of {G}eneral
  {K}nowledge}.
\newblock In \emph{The Proceedings of AAAI}, pages 4444--4451.

\bibitem[{Van Der~Walt et~al.(2011)Van Der~Walt, Colbert, and
  Varoquaux}]{van2011numpy}
Stefan Van Der~Walt, S~Chris Colbert, and Gael Varoquaux. 2011.
\newblock {T}he {N}um{P}y {A}rray: {A} {S}tructure for {E}fficient {N}umerical
  {C}omputation.
\newblock \emph{Computing in Science \& Engineering}, 13(2):22.

\bibitem[{Van~Rossum and Drake(2009)}]{10.5555/1593511}
Guido Van~Rossum and Fred~L. Drake. 2009.
\newblock \emph{{P}ython 3 {R}eference {M}anual}.
\newblock CreateSpace, Scotts Valley, CA.

\bibitem[{{Virtanen} et~al.(2020){Virtanen}, {Gommers}, {Oliphant},
  {Haberland}, {Reddy}, {Cournapeau}, {Burovski}, {Peterson}, {Weckesser},
  {Bright}, {van der Walt}, {Brett}, {Wilson}, {Jarrod Millman}, {Mayorov},
  {Nelson}, {Jones}, {Kern}, {Larson}, {Carey}, {Polat}, {Feng}, {Moore}, {Vand
  erPlas}, {Laxalde}, {Perktold}, {Cimrman}, {Henriksen}, {Quintero}, {Harris},
  {Archibald}, {Ribeiro}, {Pedregosa}, {van Mulbregt}, and
  {Contributors}}]{2020SciPy-NMeth}
Pauli {Virtanen}, Ralf {Gommers}, Travis~E. {Oliphant}, Matt {Haberland}, Tyler
  {Reddy}, David {Cournapeau}, Evgeni {Burovski}, Pearu {Peterson}, Warren
  {Weckesser}, Jonathan {Bright}, St{\'e}fan~J. {van der Walt}, Matthew
  {Brett}, Joshua {Wilson}, K.~{Jarrod Millman}, Nikolay {Mayorov}, Andrew
  R.~J. {Nelson}, Eric {Jones}, Robert {Kern}, Eric {Larson}, CJ~{Carey},
  {\.I}lhan {Polat}, Yu~{Feng}, Eric~W. {Moore}, Jake {Vand erPlas}, Denis
  {Laxalde}, Josef {Perktold}, Robert {Cimrman}, Ian {Henriksen}, E.~A.
  {Quintero}, Charles~R {Harris}, Anne~M. {Archibald}, Ant{\^o}nio~H.
  {Ribeiro}, Fabian {Pedregosa}, Paul {van Mulbregt}, and SciPy 1.~0
  {Contributors}. 2020.
\newblock \href {https://doi.org/https://doi.org/10.1038/s41592-019-0686-2}
  {{{S}ci{P}y 1.0: {F}undamental {A}lgorithms for {S}cientific {C}omputing in
  {P}ython}}.
\newblock \emph{Nature Methods}, 17:261--272.

\end{thebibliography}
\bibliographystyle{acl_natbib}

\appendix

\section{Statistical Significance}

For all claims of statistical significance we use Welch's t-test, as implemented in scipy \cite{2020SciPy-NMeth}, using an alpha value of $0.05$.

\end{document}